\documentclass[11pt,a4paper]{article}
\usepackage[hyperref]{naaclhlt2019}
\usepackage{times}
\usepackage{latexsym}
\usepackage{tabularx}
\usepackage{tabulary}

\usepackage{url}

\aclfinalcopy 

\title{Distinguishing Clinical Sentiment: The Importance of Domain Adaptation in Psychiatric Patient Health Records}

\author{Eben Holderness\textsuperscript{1,2},  Philip Cawkwell\textsuperscript{1}, Kirsten Bolton\textsuperscript{1}, \\ {\bf James Pustejovsky\textsuperscript{2}} \and {\bf Mei-Hua Hall\textsuperscript{1}} \\
\textsuperscript{1}Psychosis Neurobiology Laboratory, McLean Hospital, Harvard Medical School \\ \textsuperscript{2}Department of Computer Science, Brandeis University \\ {\tt \{eholderness, mhall\}@mclean.harvard.edu} \\ {\tt \{pcawkwell, kbolton\}@partners.org} \\ {\tt jamesp@cs.brandeis.edu}\\ }

\date{3/6/2019}

\begin{document}
\maketitle
\begin{abstract}
  Recently natural language processing (NLP) tools have been developed to identify and extract salient risk indicators in electronic health records (EHRs). Sentiment analysis, although widely used in non-medical areas for improving decision making, has been studied minimally in the clinical setting.  In this study, we undertook, to our knowledge, the first domain adaptation of sentiment analysis to psychiatric EHRs by defining psychiatric clinical sentiment, performing an annotation project, and evaluating multiple sentence-level sentiment machine learning (ML) models. Results indicate that off-the-shelf sentiment analysis tools fail in identifying clinically positive or negative polarity, and that the definition of clinical sentiment that we provide is learnable with relatively small amounts of training data. This project is an initial step towards further refining sentiment analysis methods for clinical use. Our long-term objective is to incorporate the results of this project as part of a machine learning model that predicts inpatient readmission risk. We hope that this work will initiate a discussion concerning domain adaptation of sentiment analysis to the clinical setting. 
 
\end{abstract}

\section{Introduction}

Psychotic disorders typically emerge in late adolescence or early adulthood \cite{kessler2007age, thomsen1996schizophrenia} and affect approximately 2.5-4\% of the population \cite{perala2007lifetime, bogren2009common}, making them one of the leading causes of disability worldwide \cite{vos2015global}. A substantial proportion of psychiatric inpatients are readmitted after discharge \cite{wiersma1998natural}. Readmissions are disruptive for both patients and families, and are a key driver of rising healthcare costs \cite{mangalore2007cost, wu2005economic}. Reducing readmission risk is therefore a major unmet need of psychiatric care. Developing clinically implementable ML tools to enable accurate assessment of readmission risk factors offers opportunities to inform the selection of treatment interventions and to subsequently implement appropriate preventive measures. 

Sentiment analysis (also known as opinion mining) has been used for capturing the subjective ``feeling" (e.g. positive, negative, or neutral valence) of reviews and has recently been expanded to include other domains such as reactions to stock market prediction or political trends \cite{mantyla2018evolution}. With the rise of social media and other user-generated web content, sentiment analysis has been adopted by many industries as a way of monitoring opinions towards their products, reputations, and for identifying opportunities for improvement. 
Traditionally, sentiment analysis has been approached with a lexicon-based majority vote approach, where a dictionary of terms and their associated sentiments (e.g. SentiWordnet, Pattern, SO-CAL, VADER) are queried to determine the sentiment of a given text \cite{taboada2011lexicon}. However, this approach fails to account for many associated linguistic challenges such as negation handling, scope, sarcasm, qualified statements, and out-of-vocabulary terms. As such, research groups have moved towards approaching the problem from a corpus-based machine learning perspective. This approach has the added benefit of model flexibility depending on the training data and can capture more syntactic nuance. Most state-of-the-art performances on sentiment analysis benchmarks are currently achieved with deep learning sequence models that are trained on syntactically parsed corpora such as the Stanford Sentiment Treebank \cite{socher2013recursive}.

In clinical and medical domains, however, sentiment analysis has not yet been well studied. Yet retrieving subjective clinical attitudes (sentiment) from EHR narratives has the potential to facilitate identification of a patient's symptomatological worsening or increased readmission risk.

The concept of medical sentiment is complex and vocabularies differ from general-domain sentiment. In the field of psychiatry, this is especially true. Therefore, there is a need for domain adaptation of sentiment analysis that includes a richer array of attributes than can typically be found in off-the-shelf tools. In this work, we established an annotation scheme to characterize sentiment-related features in EHRs, and used this to carry out, to our knowledge, the first psychiatry-specific sentiment annotation project on EHRs. The resulting datasets are used to train and evaluate a classifier to predict clinical sentiment at the sentence level. This classifier, which in future works will be integrated in a pipeline for predicting readmission risk, is clinically useful for targeting treatments and aiding in decision making.

\begin{table*}
\centering
\small
\begin{tabulary}{0.94\linewidth}{|p{1.5cm}|p{4.25cm}|p{4.25cm}|p{4.25cm}|}
\hline
{\bf Domain} & {\bf  Positive Example} & {\bf Neutral Example} & {\bf Negative Example} \\\hline
Appearance & Presents on time, dressed and groomed nicely, good hygiene. & Casually dressed and wearing knit vest and belt. & Notes that he wears the same clothes 2-3 days at a time, he doesn't care for his appearance--which is atypical for him.\\\hline 
Mood & Her depression and anxiety have improved immensely. & Mood is largely euthymic although he stated he gets depressed occasionally. & Tearful, presented very depressed with sad affect.\\\hline
Interpersonal & Continues to be happy in her relationship with her boyfriend and school friends are stable as well. & She voiced no complaints about her primary relationship or other social relationships. & Poor social supports, abusive relationship.\\\hline
Substance Use & Denies substance use or alcohol other than an occasional glass of wine. & Remote history of cocaine (smoked), marijuana and mescaline use many years ago. & He reports daily k2 use in addition to using crack cocaine about once a week. \\\hline
Occupation & Pt reports having taken further steps toward employment -- applied for two jobs and has interview lined up for Saturday. & Discusses new job as part time substitute teacher. & Recently has a new job that she hates and took a paycut.\\\hline
Thought Content & She never had auditory hallucinations or delusions of thought broadcasting and thought insertion. &  No overt hallucinations or delusions but expansive thinking. &  Delusions and hallucinations continue.\\\hline 
Thought Process & Stable, slow speech with fewer word finding difficulties today, linear thought process, cooperative,attentive. & Slightly pressured speech but not as bad as some past visits. & Speech spontaneous and decreased in volume, rate, and rhythm; hard to understand at times because she barely opens her mouth when she talks.
\\\hline
\end{tabulary}
\caption{Example EHR sentences reflecting sentiment polarity for each risk factor domain.}\label{example-sents}
\end{table*}

\section{Related Works}
Although there has been some work on clinical adaptation of sentiment analysis using healthcare-related data extracted from web forums, biomedical texts, or social media postings (See for example \cite{smith2012cross, niu2005analysis, salas2017sentiment, nguyen2014affective}), there has been minimal work on sentiment analysis when applied to actual EHR data. 

McCoy et al. \shortcite{mccoy2015sentiment} used a corpus of psychosis patient discharge summaries and the 3,000 word Pattern lexical opinion mining dictionary \cite{smedt2012pattern} to classify the associated sentiment of documents using a majority vote classifier. Results of their Cox regression models showed that greater positive sentiment was associated with a reduction in inpatient readmission risk. Waudby-Smith et al. \shortcite{waudby2018sentiment} applied the same Pattern sentiment lexicon to a corpus of ICU nursing notes to predict 30-day mortality risk. They found that stronger negative sentiment polarity was associated with an increased 30-day mortality risk. One of the limitations in both studies is that Pattern is a general-domain sentiment lexicon that contains few informative medical or psychiatry-specific terminology. Also, the authors did not manually annotate the datasets they worked with. As a result, they were not able to confirm that the predicted sentiment aligned with the sentiment from a clinical perspective. 

\cite{deng2014retrieving} and \cite{denecke2015sentiment} systematically compared word usage and sentiment distribution between clinical narratives (nurse letters, discharge summary, and radiology reports) and medical social media (MedBlog, drug reviews). They concluded that off-the-shelf sentiment tools were not ideal for analyzing sentiment in medical documents and that EHRs were significantly more difficult in predicting sentiment, in particular neutral sentiment (Neutral F1=0.216 and 0.080 for nurse letters and radiology reports, respectively). They developed annotation guidelines and undertook a span-level annotation task on 300 ICU nurse letters to identify words related to clinical sentiment \cite{deng2016generation}. Results of applying ML algorithms to these data are not available yet.

\section{Methods}

In this work, we define psychiatric clinical sentiment as a clinician's attitudes (positive, negative, or neutral) towards a patient's prognosis with regards to seven readmission risk factor domains (appearance, mood, interpersonal relations, substance use, thought content, thought process, and occupation) that were identified in prior work \cite{holderness2018analysis}. The scope of our current definition is intentionally narrow such that the sentiment of a given sentence is considered in isolation without any prior knowledge.

Three clinicians participated in an annotation project that focused on identifying the clinical sentiment associated with psychiatric EHR texts at the sentence level. In total, two corpora of clinical narratives from institutional EHRs, one containing 3,500 sentences (training dataset) and the other 1,650 (test dataset) were annotated using the definition established in the annotation scheme. 

The training dataset consisted exclusively of sentence-length sequences that involved only one risk factor domain in each example. The examples in the dataset were identified from a large corpus of unannotated psychosis patient EHR data sourced from the psychiatric units of several Boston-area hospitals in the Partners HealthCare network, including Massachusetts General Hospital and Brigham \& Women's Hospital. We used our risk factor domain topic extraction model to automatically identify relevant sentences, which were then manually validated by one of the clinicians involved in this project to ensure they did not involve multiple domains in the same example. See Table \ref{example-sents} for example sentences for each domain. 

The test dataset is an extension of the corpus used previously to evaluate our risk factor domain topic extraction model and is non-overlapping with the training data, consisting of discharge summaries, admission notes, individual encounter notes, and other clinical notes from 220 patients in the OnTrack\textsuperscript{TM} program at McLean Hospital. OnTrack\textsuperscript{TM} is an outpatient program, focusing on treating adults ages 18 to 30 who are experiencing their first episodes of psychosis. Because we are interested in identifying the clinical sentiment associated with each risk factor domain individually, the test dataset consists of examples that were intentionally selected to be challenging for our model: they are variable in length, wide-ranging in vocabulary, and can involve multiple risk factor domains (e.g. ``Work functioning is impaired, but pt has good relationship w/ his girlfriend and is not engaging in substance use."). 

These corpora are available to other researchers upon request. Table \ref{distributions} details the distribution of the training and test data. The imbalance of training examples across the three sentiment classes reflects the natural distribution of sentiment reflected in EHRs, as certain risk factor domains (e.g. substance use) will rarely be reflected in a neutral or positive sense. 

\begin{table}[]
\centering
\small
\begin{tabular}{llll}
 & \textbf{Positive} & \textbf{Negative} & \textbf{Neutral} \\
\textbf{Appearance} & 290 & 69 & 141 \\
\textbf{Mood} & 100 & 322 & 77 \\
\textbf{Interpersonal} & 205 & 165 & 130 \\
\textbf{Substance Use} & 181 & 261 & 58 \\
\textbf{Occupation} & 250 & 143 & 150 \\
\textbf{Thought Process} & 150 & 266 & 84 \\
\textbf{Thought Content} & 183 & 253 & 64
\end{tabular}
\caption{Distribution of training and test examples.}
\label{distributions}
\end{table}

\tabcolsep=1pt\relax
\begin{table*}[]
\begin{footnotesize}
\centering
\begin{tabular*}{\textwidth}{lllllllllll}
\textbf{Model} & \textbf{Domain} & \textbf{Pos P} & \textbf{Pos R} & \textbf{Pos F1} & \textbf{Neg P} & \textbf{Neg R} & \textbf{Neg F1} & \textbf{Neu P} & \textbf{Neu R} & \textbf{Neu F1} \\
\textbf{Baseline (Pattern)} & All & 0.612 & 0.231 & 0.319 & 0.552 & 0.245 & 0.337 & 0.234 & \textbf{0.736} & \textbf{0.348} \\
 & Interpersonal & 0.8 & 0.222 & 0.348 & 0.429 & 0.103 & 0.167 & 0.413 & 0.929 & 0.571 \\
 & Mood & 0.511 & 0.233 & 0.32 & 0.558 & 0.352 & 0.432 & 0.266 & 0.672 & 0.381 \\
 & Occupation & 0.75 & 0.129 & 0.22 & 0.328 & 0.188 & 0.265 & 0.329 & 0.917 & 0.484 \\
 & Substance Use & 0.429 & 0.067 & 0.115 & 0.593 & 0.241 & 0.342 & 0.222 & 0.74 & 0.341 \\
 & Appearance & 0.781 & 0.424 & 0.549 & 0.556 & 0.309 & 0.397 & 0.174 & 0.552 & 0.265 \\
 & Thought Content & 0.556 & 0.19 & 0.283 & 0.723 & 0.29 & 0.414 & 0.055 & 0.6 & 0.101 \\
 & Thought Process & 0.459 & 0.354 & 0.4 & 0.677 & 0.231 & 0.344 & 0.181 & 0.739 & 0.291 \\
\textbf{Fully Supervised MLP} & All & \textbf{0.62} & 0.416 & \textbf{0.478} & \textbf{0.67} & 0.652 & \textbf{0.658} & \textbf{0.289} & 0.437 & 0.329 \\
 & Interpersonal & 0.632 & 0.667 & 0.649 & 0.731 & 0.656 & 0.691 & 0.567 & 0.607 & 0.558 \\
 & Mood & 0.717 & 0.32 & 0.443 & 0.597 & 0.73 & 0.657 & 0.286 & 0.418 & 0.339 \\
 & Occupation & 0.645 & 0.571 & 0.606 & 0.558 & 0.604 & 0.58 & 0.346 & 0.375 & 0.36 \\
 & Substance Use & 0.423 & 0.244 & 0.31 & 0.674 & 0.714 & 0.693 & 0.344 & 0.42 & 0.378 \\
 & Appearance & 0.705 & 0.525 & 0.602 & 0.69 & 0.605 & 0.645 & 0.241 & 0.448 & 0.313 \\
 & Thought Content & 0.59 & 0.127 & 0.209 & 0.667 & 0.654 & 0.66 & 0.078 & 0.4 & 0.13 \\
 & Thought Process & 0.629 & 0.458 & 0.53 & 0.775 & 0.604 & 0.679 & 0.161 & 0.391 & 0.228 \\
\textbf{\begin{tabular}[c]{@{}l@{}}Semi-Supervised MLP \\ (Self-Training)\end{tabular}} & All & 0.588 & 0.4 & 0.46 & 0.611 & \textbf{0.733} & \textbf{0.658} & 0.285 & 0.291 & 0.259 \\
 & Interpersonal & 0.632 & 0.667 & 0.649 & 0.625 & 0.69 & 0.656 & 0.583 & 0.5 & 0.539 \\
 & Mood & 0.645 & 0.301 & 0.411 & 0.502 & 0.885 & 0.641 & 0.233 & 0.105 & 0.144 \\
 & Occupation & 0.671 & 0.671 & 0.671 & 0.539 & 0.583 & 0.56 & 0.364 & 0.333 & 0.348 \\
 & Substance Use & 0.394 & 0.289 & 0.333 & 0.617 & 0.835 & 0.709 & 0.333 & 0.1 & 0.154 \\
 & Appearance & 0.722 & 0.441 & 0.547 & 0.653 & 0.605 & 0.628 & 0.224 & 0.448 & 0.299 \\
 & Thought Content & 0.5 & 0.139 & 0.218 & 0.689 & 0.753 & 0.72 & 0.088 & 0.333 & 0.139 \\
 & Thought Process & 0.583 & 0.292 & 0.389 & 0.651 & 0.78 & 0.685 & 0.172 & 0.217 & 0.192 \\
\end{tabular*}
\caption{Results of the clinical sentiment extraction task.}
\label{results}
\end{footnotesize}
\end{table*}

We evaluated three classification models. Our baseline model is a majority vote approach using the Pattern sentiment lexicon employed by McCoy \shortcite{mccoy2015sentiment} and Waudby-Smith \shortcite{waudby2018sentiment}. The second and third models use fully supervised and semi-supervised multilayer perceptron (MLP) architectures, respectively. Since positive and negative clinical sentiment can differ across each domain, we train a suite of seven models, one for each risk factor domain. The training and test data were vectorized at the sentence level using the pretrained Universal Sentence Encoder (USE) embedding module \cite{cer2018universal} that is available through TensorFlow Hub and is designed specifically for transfer learning tasks. Although USE is trained on a large volume of web-based, general-domain data, we have found in prior work that the embeddings lead to higher accuracy on downstream classification tasks than embedding models (e.g. ELMo, Doc2Vec, FastText) trained on smaller volumes of EHR data \cite{Me2019}. 

Hyperparameters were tuned using grid search with 5-fold cross-validation on the training dataset and are specified in Table \ref{hyperparameters}. Due to the relatively small amount of labeled training data, our proposed model architecture is designed to prevent overfitting by using a restricted view of the training data via a high rate of dropout in the hidden layers. Additionally, we use two hidden layers to extract a more abstracted form of the input. Additionally, because neutral sentiment is much broader in scope and has fewer training examples, resulting in covariate shift, we compute a threshold for classifying positive and negative sentiment using the formula min=avg(sim)+α*σ(sim), where σ is standard deviation and α is a constant, which we set to 0.2. If a given test sentence does not have positive or negative outputs that exceed this threshold, the sentence is classified as neutral even if neutral is not the maximal output. 

\begin{table}[]
\centering
\small
\begin{tabular}{ll}
\textbf{Parameter} & \textbf{Value} \\
Batch Size & 28 \\
Iterations & 100 \\
Hidden Units Per Layer & 300 \\
Dropout & 0.75 \\
Kernel Initializer & Uniform \\
Optimizer & Adam \\
Input/Hidden Layer Activations & ReLU \\
Output Layer Activations & Sigmoid
\end{tabular}
\caption{Hyperparameters for sentiment model.}
\label{hyperparameters}
\end{table}

We experimented with two semi-supervised learning configurations, Self-Training and K-Nearest Neighbors (KNN). The self-training approach involved first training our model on the labeled training data and then using this model to identify unlabeled examples from a large preprocessed corpus of unlabeled EHR data (2,100,000 sentences, 85,000,000 tokens). For the KNN approach, we projected all of the labeled and unlabeled examples into vector space and treated the labeled examples as centroids. For each centroid, we then used Euclidean distance to compute the five nearest unlabeled examples. Both models were trained using a 20:80 combination of the original labeled data and the additional unlabeled data. 

\section{Results and Discussion}

Inter-annotator agreement (IAA) was substantial on the first corpus (Scott's Pi=0.691, Cohen's Kappa=0.693) and higher on the second (Pi=0.768, Kappa=0.768) \cite{fleiss1971measuring,davies1982measuring}. This is expected as the first corpus contains many sentences involving multiple readmission risk factor domains and annotators were instructed to provide clinical sentiment labels for each, whereas the second corpus consists entirely of single domain sentences. In both cases, IAA surpasses that reported by Denecke and Deng \shortcite{deng2016generation}, primarily because of the clinical expertise of the annotators involved in this project.

Results of the three classifiers are shown in Table \ref{results}, with the highest score on each performance  metric in bold. The `All' row for each model configuration was computed by averaging the scores of the sentiment models for each risk factor domain. Applying the Pattern sentiment lexicon to our test corpus showed a strong trend towards underclassification of positive and negative examples, which led to poor recall scores while maintaining moderate precision. Neutral examples, however, were correctly classified significantly more often. This confirms that many of the most informative words in terms of clinical sentiment (e.g. `hallucination', `depressed', `employed', etc.) do not hold significance in general-domain sentiment and are therefore not part of the Pattern lexicon. 

Despite the relatively small size of the training corpus, the EHR data used for training captured much of the domain-specific vocabulary related to clinical sentiment and our suite of models achieved F-measures on classifying positive and negative sentiment that exceed those reported in prior literature \cite{deng2014retrieving}. Although direct comparison between our EHR dataset and the EHR datasets used by other researchers is limited due to HIPAA restrictions, our training EHR data is sourced from the same EHR database as McCoy \shortcite{mccoy2015sentiment}. Therefore, a better performance of our models indirectly supports that our model can better capture the underlying clinical sentiment embedded in EHRs.

Because clinical documents are written for a specific purpose such as assessing the outcome of treatment, they contain less neutral content and as a result sentiment distributions are intrinsically biased to either positive or negative polarity. Thus, identifying training examples with neutral sentiment was challenging and consequently both the fully and semi-supervised models were poor at identifying neutral sentiment across all seven domains. In addition, unless the patient is markedly improved, clinicians tend to document continuing unresolved symptoms. leading to a greater amount of negative content. We hypothesize that this may be one reason for the lower overall F1 performance on positive versus negative sentiment. 

We observed that per-domain performance of our models aligned with the natural distribution of positive vs. negative clinical sentiment in EHRs. Substance use, for example, had low positive F1 scores as the majority of references to substance use in EHRs involve negative sentiment unless the patient is noted to be abstaining from substance use. We also observed that sentiment distribution towards negative polarity is more evident in mood and thought content, which include, for example, delusions, depression, anxiety, and hallucinations. 

When applying semi-supervised learning methods, we found self-training to marginally improve performance on negative clinical sentiment but the overall F1 score was not better than the fully
supervised model due to lower precision. We observed minimal change in performance when using a k-nearest neighbors approach.  

\section{Conclusion and Future Work}

We focused in this study on the clinical sentiment associated with readmission for seven risk factor domains identified in prior work by undertaking an annotation project and using the resultant gold standard to train semi-supervised ML algorithms to automatically infer this sentiment. Our results indicate that domain adaptation of sentiment analysis is necessary for aligning with clinician opinions. 

We intend to improve our clinical sentiment classifier in future work by increasing the size of the annotated training corpus (in particular neutral examples) and by changing the model input to a sequence model as opposed to a full sentence vector representation. We also intend to modify our definition of clinical sentiment to include temporal linking of elements that involve clinical sentiment in an EHR to establish gradients of changes in patient status over time. Finally, we will incorporate our sentiment analysis model in a classifier that predicts inpatient readmission risk.

\section{Acknowledgments}
This work was supported by a grant from the National Institute of Mental Health (grant no. 5R01MH109687 to Mei-Hua Hall). We would also like to thank the Clinical NLP 2019 Workshop reviewers for their constructive and helpful comments. 

\bibliography{naaclhlt2019}
\bibliographystyle{acl_natbib}
\end{document}